\documentclass[journal]{IEEEtran}

\usepackage{bobbystyle_IEEEjournal}

\usepackage{tikz}
\usetikzlibrary{spy}

\newcommand{\dquotes}[1]{``#1''}

\usepackage{chngcntr}
\counterwithout{equation}{section}

\tikzstyle{only in spy node magn 1.75}=[%
   transform canvas={%
      shift={(tikzspyinnode)},
      scale=1.75,
   }
]

\usepackage{balance}

\newcolumntype{C}[1]{>{\centering\let\newline\\\arraybackslash\hspace{0pt}}m{#1}}

  {\list{}{\leftmargin=#1\rightmargin=#1}\item[]}%
  {\endlist}

\newcommand{\Ds}{D^\star}
\newcommand{\Dt}{D_{\textnormal{true}}}

\newcommand{\Zt}[1]{Z_{\textnormal{true},#1}}
\newcommand{\ZZ}{Z}
\newcommand{\Z}[1]{Z_{#1}}
\newcommand{\eig}[1]{\lambda_{#1}}
\newcommand{\eigmin}{\eig{\textnormal{min}}}
\newcommand{\eigmax}{\eig{\textnormal{max}}}
\newcommand{\svd}[1]{\sigma_{#1}}
\newcommand{\Lambdab}{\bar{\Lambda}}
\newcommand{\Q}{\mbox{Q}}
\newcommand{\vect}{\operatorname{vec}}
\newcommand{\sigmab}{\bar{\sigma}}
\newcommand{\rhob}{\bar{\rho}}
\newcommand{\chib}{\bar{\chi}}

\begin{document}

\title{\fontsize{23.4}{23.4}\selectfont Convolutional Analysis Operator Learning:\\ Dependence on Training Data}

\urldef{\chunmail}\url{iychun@umich.edu}
\urldef{\hongmail}\url{dahong@umich.edu}
\urldef{\fesslermail}\url{fessler@umich.edu}
\urldef{\adcockmail}\url{ben_adcock@sfu.ca}

\author{Il Yong Chun,$^{\dagger}$ \textit{Member}, \textit{IEEE}, David Hong,$^{\dagger}$ \textit{Student Member}, \textit{IEEE}, \\ Ben Adcock, and Jeffrey A. Fessler, \textit{Fellow}, \textit{IEEE}

\vspace{-1.2\baselineskip}

\thanks{
$\dagger$\textit{The first two authors contributed equally to this work.}
}

\thanks{This work is supported in part by the Keck Foundation and NIH grant U01 EB018753. BA is supported by NSERC grant 611675.}

\thanks{Il Yong Chun, David Hong, and Jeffrey A. Fessler are with the Department of Electrical Engineering and Computer Science, The University of Michigan, Ann Arbor, MI 48019 USA (email: \chunmail; \hongmail; \fesslermail). Ben Adcock is with the Department of Mathematics, Simon Fraser University, Burnaby, BC V5A 1S6 Canada (email: \adcockmail)}
}

\maketitle
\begin{abstract}
Convolutional analysis operator learning (CAOL) enables the unsupervised training of (hierarchical) convolutional sparsifying operators or autoencoders from large datasets.
One can use many training images for CAOL, but a precise understanding of the impact of doing so has remained an open question.
This paper presents a series of results that lend insight into the impact of dataset size on the filter update in CAOL.
The first result is a general deterministic bound on errors in the estimated filters, and is followed by a bound on the {\em expected} errors as the number of training samples increases.
The second result provides a {\em high probability} analogue.
The bounds depend on properties of the training data, and we investigate their empirical values with real data.
Taken together, these results provide evidence for the potential benefit of using more training data in CAOL.
\end{abstract}

\section{Introduction}

\IEEEPARstart{L}{earning} convolutional operators from large datasets is a growing trend in signal/image processing, computer vision, machine learning, and artificial intelligence.
The \emph{convolutional} approach resolves the large memory demands of patch-based operator learning and enables unsupervised operator learning from \dquotes{big data,} i.e., many high-dimensional signals.
See \cite{Chun&Fessler:18arXiv, Chun&Fessler:18TIP} and references therein.
Examples include \emph{convolutional dictionary learning} \cite{Chun&Fessler:18TIP, Chun&Fessler:17SAMPTA} and \emph{convolutional analysis operator learning} (CAOL) \cite{Chun&Fessler:18arXiv, Chun&Fessler:18Asilomar}.
CAOL trains an autoencoding convolutional neural network (CNN) in an unsupervised manner, and is useful for training multi-layer CNNs \cite{Chun&Fessler:18arXiv} and iterative CNNs \cite{Chun&etal:18arXiv:momnet, Chun&Fessler:18IVMSP, Chun&etal:18Allerton} from many training images.
In particular, the block proximal extrapolated gradient method using a majorizer \cite{Chun&Fessler:18TIP, Chun&Fessler:18arXiv} leads to rapidly converging and memory-efficient CAOL \cite{Chun&Fessler:18arXiv}.
However, a theoretical understanding of the impact of using many training images in CAOL has remained an open question.

This paper presents new insights on this topic.
Our first main result provides a {\em deterministic} bound on filter estimation error, and is followed by a bound on the {\em expected} error when ``model mismatch'' has zero mean.
(See Theorem~\ref{t:det} and Corollary \ref{c:det:randNoise}, respectively.)
The expected error bound depends on the training data, and we provide empirical evidence of its decrease with an increase in training samples.
Our second main result provides a {\em high probability} bound that explicitly decreases with increasingly many i.i.d. training samples. The bound improves when model mismatch and samples are uncorrelated.
(See Theorem~\ref{t:conc}.)
Additional empirical findings provide evidence that the correlation can indeed be small in practice.
Put together, our findings provide new insight into how using many  samples can improve CAOL, 
underscoring the benefits of the low memory usage of CAOL.

\section{Backgrounds and Preliminaries}

\subsection{CAOL with orthogonality constraints} \label{sec:model}

CAOL seeks a set of filters that \dquotes{best} sparsify a set of training images $\{ x_l \in \bbC^{N} : l = 1,\ldots,L \}$ by solving the optimization problem \cite[\S\Romnum{2}-A]{Chun&Fessler:18arXiv} (see Appendix for notation):
\begingroup
\setlength{\thinmuskip}{1.5mu}
\setlength{\medmuskip}{2mu plus 1mu minus 2mu}
\setlength{\thickmuskip}{2.5mu plus 2.5mu}
\fontsize{9.5pt}{11.4pt}\selectfont
\ea{
\label{sys:CAOL:orth}
\argmin_{D = [d_1, \ldots, d_K]} \min_{\{ z_{l,k} \}} &~ F(D, \{ z_{l,k} \}), \quad \mbox{subj.~to}~ D D^H = \frac{1}{R} \cdot I, \tag{P0}
\\
F(D, \{ z_{l,k} \}) & :=  \sum_{l=1}^L \sum_{k=1}^K  \left\| d_k \circledast x_l - z_{l,k} \right\|_2^2 + \alpha \| z_{l,k} \|_0, \nn
}
\endgroup
where $\circledast$ denotes convolution, $\{ d_k \in \bbC^{R} : k = 1,\ldots,K\}$ is a set of $K \geq R$ convolutional kernels, $\{ z_{l,k} \in \bbC^{N} : l = 1,\ldots,L, k = 1,\ldots,K \}$ is a set of sparse codes,
$\alpha \!>\! 0$ is a regularization parameter controlling the sparsity of features $\{  z_{l,k} \}$,
and $\nm{\cdot}_{0}$ denotes the $\ell^0$-quasi-norm.
We group the $K$ filters into a matrix:
\be{
\label{eq:D}
D := \left[ \arraycolsep=2pt \begin{array}{ccc} d_1 & \cdots & d_K \end{array} \right]
\in \bbC^{R \times K}
.
}
The orthogonality condition $D D^H = \frac{1}{R} I$ in \R{sys:CAOL:orth} enforces \emph{1)} a tight-frame condition on the filters, i.e., $\sum_{k=1}^K \nm{ d_k \circledast x }_2^2 = \nm{ x }_2^2$, $\forall x$ \cite[Prop.~2.1]{Chun&Fessler:18arXiv};
and \emph{2)} filter diversity when $R=K$, since $D D^H = \frac{1}{R} I$ implies $D^H D = \frac{1}{K} I$ and each pair of filters is incoherent, i.e., $| \ip{d_k}{d_{k'}} |^2 = 0$, $\forall k \neq k'$.
One often solves \R{sys:CAOL:orth} iteratively, by alternating between optimizing $D$ (filter update) and optimizing $\{ z_{l,k} : \forall l, k \}$ (sparse code update) \cite{Chun&Fessler:18arXiv}, i.e., at the $i$ iteration, the current iterates are updated as
$\{ z_{l,k}^{(i+1)} \} = \argmin_{\{ z_{l,k} \}} F(D^{(i)}, \{ z_{l,k} \})$ and $D^{(i+1)} = \argmin_{D D^H = \frac{1}{R} \cdot I} F(D, \{ z_{l,k}^{(i+1)} \})$.

\subsection{Filter update in a matrix form}

The key to our analysis lies in rewriting the filter update for \R{sys:CAOL:orth} in matrix form, to which we apply matrix perturbation and concentration inequalities.
Observe first that
\be{
\label{eq:Psi_l}
d_k \circledast x_l
=
\underbrace{
  [\Pi^0 x_l, \dots, \Pi^{R-1} x_l]
}_{\mbox{$=:\Psi_l \in \bbC^{N \times R}$}}
d_k
= \Psi_l d_k
,
\quad l = 1,\ldots,L,
}
where $\Pi := \left[
\begin{smallmatrix}
0 & I_{N-1} \\ 1 & 0
\end{smallmatrix} \right]
\in \bbC^{N \times N}
$
is the circular shift operator and
$(\cdot)^n$ denotes the matrix product of its $n$ copies.
We consider a circular boundary condition to simplify the presentation of $\{ \Psi_l \}$ in \R{eq:Psi_l}, but our entire analysis holds for a general boundary condition with only minor modifications of $\{ \Psi_l \}$ as done in \cite[\S\Romnum{4}-A]{Chun&Fessler:18arXiv}.
Using \R{eq:Psi_l}, the filter update of \R{sys:CAOL:orth} is rewritten as
\be{
\label{opt:filter:orth}
D^\star = \argmin_{D} \sum_{l=1}^L \left\| \Psi_l D - Z_l \right\|_F^2,  \quad \mbox{subj.~to} ~ D D^H = \frac{1}{R} \cdot I,
\tag{P1}
}
where $Z_l := [ z_{l,1}, \ldots, z_{l,K} ] \in \bbC^{{N} \times K}$ contains all the current sparse code estimates for the $l$th sample, and we drop iteration superscript indices $(\cdot)^{(i)}$ throughout.
The next section uses this form to characterize the filter update solution $D^\star$.

\section{Main Results: \\ Dependence of CAOL on Training Data}
\label{sec:caol:data}

The main results in this section illustrate how training with many samples can reduce errors in the filter $\Ds$ from \R{opt:filter:orth} and characterize the reduction in terms of properties of the training data.
Throughout we model the current sparse codes estimates as
\be{
\label{sys:filter}
Z_l
=
\underbrace{ \Psi_l \Dt }_{\mbox{$=: \Zt{l}$}} + E_l,
\quad l = 1,\dots,L
}
where $\Dt$ is formed from optimal (orthogonal) filters analogously to \R{eq:D},
and $E_l \in \bbC^{N \times K}$ captures model mismatch in the current sparse codes,
e.g., due to the current iterate being far from convergence or being trapped in local minima.

The following theorem provides a {\em deterministic} characterization.

\thm{
\label{t:det}
Suppose that both matrices
\be{
\label{t:det:bound:rowrank}
\sum_{l=1}^L \Psi_l^H \Z{l}  \in \bbC^{R \times K}
\quad\mbox{and}\quad
\sum_{l=1}^L \Psi_l^H \Zt{l} \in \bbC^{R \times K}
}
are full row rank, where $\{ \Psi_l, \Z{l}, \Zt{l} : l=1,\ldots,L \}$
are defined in \R{eq:Psi_l}--\R{sys:filter}.
Then, the solution $\Ds$ to \R{opt:filter:orth} has error with respect to $\Dt$ bounded as
\be{
\label{t:det:bound:new}
\| \Ds - \Dt \|_F^2
\leq 5
\frac{ \| \sum_{l=1}^L \Psi_l^H E_l \|_F^2 }
  { \eigmin^2( \sum_{l=1}^L \Psi_l^H \Psi_l ) },
}
where $\eigmin(\cdot)$ denotes the smallest eigenvalue of its argument.
}

The full row rank condition on \R{t:det:bound:rowrank} ensures that the estimated filters $\Ds$ and the true filters $\Dt$ are unique, and it further guarantees that the denominator of \R{t:det:bound:new} is strictly positive.
When the model mismatches $E_1,\dots,E_L$ are independent and mean zero, we obtain the following expected error bound:
\cor{
\label{c:det:randNoise}
Under the construction of Theorem~\ref{t:det}, suppose that $E_l$ is a zero-mean random matrix for $l=1,\ldots,L$, and is independent over $l$.
Then,
\be{
\label{c:det:randNoise:bound}
\bbE \| \Ds - \Dt \|_F^2
\leq
5 \sigmab^2 \rho^2,
}
where $\bbE(\cdot)$ denotes the expectation,
\ea{
\label{c:det:randNoise:rho2}
\sigmab^2 &:= \max_{l=1,\ldots,L} \eigmax(\bbE\{ E_l E_l^H\}),
\nn \\
\rho^2 &:=
\frac{ \tr (\sum_{l=1}^L\Psi_l^H\Psi_l) }
  { \eigmin^2 (\sum_{l=1}^L \Psi_l^H \Psi_l) }
,
}
$\eigmax(\cdot)$ denotes the largest eigenvalue of its argument,
and the expectation is taken over the model mismatch.
}

\begin{figure}[!tp]
\centering

\begin{tabular}{c}
\includegraphics[scale=0.525, trim=0 0 0em 0em, clip]{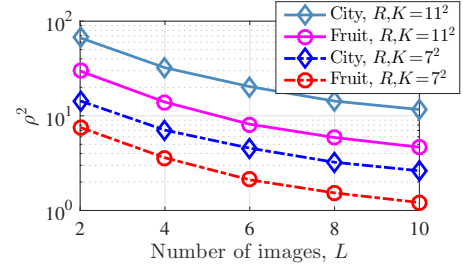}
\end{tabular}

\vspace{-0.5em}
\caption{
Empirical values of $\rho^2$ in \R{c:det:randNoise:rho2} show a decrease with $L$ for different datasets and filter dimensions. (The fruit and city datasets with $L \!=\! 10$ and $N \!=\! 10^4$ were preprocessed with contrast enhancement and mean subtraction; see details of datasets and experiments in \cite{Chun&Fessler:18arXiv, Chun&Fessler:18TIP} and references therein. For $L < 10$, the results are averaged over $50$ datasets randomly selected from the full datasets.)
Under the assumptions of Corollary \ref{c:det:randNoise}, the decrease in this quantity leads to a better expected error bound in \R{t:det:bound:new}.
Without preprocessing, the quantity $\rho^2$ increases by a factor of around $10^3$.
}
\label{fig:CAOL:data}
\end{figure}

Given fixed $K$ and $R$,
it is natural to expect that $\sigmab^2$ is bounded by some constant independent of $L$,
and so the expected error bound in \R{c:det:randNoise:bound} largely depends on $\rho^2$ in \R{c:det:randNoise:rho2}.
When training samples are i.i.d., one may further expect $(1/L) \sum_{l=1}^L \Psi_l^H \Psi_l$ to concentrate around its expectation, roughly resulting in $\rho^2 \propto 1/L$,
with a proportionality constant that depends on $R$
and the statistics of the training data.
Fig.~\ref{fig:CAOL:data} illustrates $\rho^2$ for various image datasets, providing empirical evidence of this decrease in real data.

\begin{figure}[!tp]
\centering

\begin{tabular}{c}
\includegraphics[scale=0.525, trim=0 0 0 0, clip]{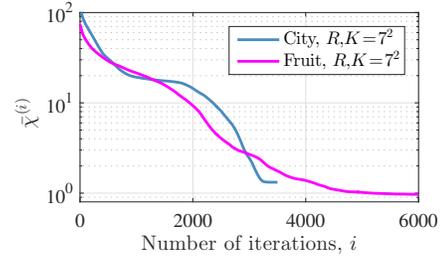}
\end{tabular}

\vspace{-0.5em}
\caption{
Empirical estimate of $\chib$ in \R{eq:conc:terms} across iterations in the alternating optimization algorithm \cite{Chun&Fessler:18arXiv} that solves CAOL~\R{sys:CAOL:orth} with $\alpha \!=\! 10^{-3}$. (The fruit and city datasets with $L \!=\! 10$ and $N \!=\! 10^4$ were preprocessed with contrast enhancement and mean subtraction; see details of datasets and experiments in \cite{Chun&Fessler:18arXiv, Chun&Fessler:18TIP} and references therein.
The model mismatches $\{ E_l^{(i)} : \forall l \}$ at the $i$th iteration were calculated every $50$ iterations based on \R{sys:filter}, where we use the converged filters for $\Dt$.)
Observe that $\chib^{(i)}$ generally decreases over iterations; when $\chib$ is small, the high probability error bound \R{t:det:bound:prob} in Theorem \ref{t:conc} depends primarily on $\rhob$ defined in \R{eq:conc:terms}.
}
\label{fig:CAOL:anal}
\end{figure}

Our second theorem provides a {\em probabilistic} error bound via concentration inequalities,
given i.i.d. training sample and model mismatch pairs
$(x_1, E_1), \dots, (x_L, E_L)$.\footnote{
We follow the natural convention in sample size analyses of assuming that $\{ x_l \!:\! \forall l \}$ are i.i.d. samples from an underlying training distribution; 
see the references cited in Section~\ref{sec:rel} and \cite{Hastie&Tibshirani&Friedman:book, Mohri&Rostamizadeh&Talwalkar:book}
for other examples.
Model mismatches $\{ E_l \!:\! \forall l \}$ also become i.i.d. across samples at all iterations of CAOL, if \dquotes{fresh} training samples are used for each update, e.g., as can be done when solving \eqref{opt:filter:orth} via mini-batch stochastic optimization.
}
It removes the zero-mean assumption for the model mismatches $\{ E_l : \forall l \}$ in Corollary~\ref{c:det:randNoise} that
might be strong, e.g., if training data are not preprocessed to have zero mean.

\thm{
\label{t:conc}
Suppose that training sample and model mismatch pairs
$(x_1, E_1), \dots, (x_L, E_L) \overset{iid}{\sim} (x,E)$,
where $x$ and $E$ are almost surely bounded, i.e.,
\be{
\label{eq:conc:asbound}
\|x\|_2 \leq \gamma \quad\mbox{and}\quad \|E\|_F \leq \sigma,
}
and the matrices in \R{t:det:bound:rowrank} are almost surely full row rank.
Then, for any $0 < \delta < \eigmin(\Lambdab)/(2R\gamma^2)$, the solution $\Ds$ to \R{opt:filter:orth} has error with respect to $\Dt$ bounded as
\begingroup
\setlength{\thinmuskip}{1.5mu}
\setlength{\medmuskip}{2mu plus 1mu minus 2mu}
\setlength{\thickmuskip}{2.5mu plus 2.5mu}
\ea{
\label{t:det:bound:prob}
&~\| \Ds - \Dt \|_F^2
\nn \\
&\leq 5
\bigg\{
\frac{
  \sigma\sqrt{\tr(\Lambdab)/L}
  +
  \| \bbE(\Psi^H E) \|_F
  +
  2 \sigma \gamma \sqrt{R} \delta
}{
  \eigmin(\Lambdab) - 2\gamma^2R \delta
}
\bigg\}^{\!\!2}
,
}
\endgroup
with probability at least
\be{
\label{t:det:prob}
1 - 3 R \exp\left(-L\frac{\delta^2/2}{3+\delta/3}\right)
,
}
where $\Lambdab := \bbE (\Psi^H \Psi)$ and $\Psi$ is constructed from $x$ as in \R{eq:Psi_l}.
}

Taking $\delta$ sufficiently small, the high probability error bound \R{t:det:bound:prob} is primarily driven by
\be{
\label{eq:conc:terms}
\rhob := \frac{\sqrt{\tr(\Lambdab)/L}}{\eigmin(\Lambdab)}
\quad \text{and} \quad
\chib := \frac{\| \bbE(\Psi^H E) \|_F}{\eigmin(\Lambdab)},
}
where $\rhob$ is analogous to $\rho$ in \R{c:det:randNoise:rho2},
and $\chib$ captures how correlated the model mismatch is to the training samples.
As the number $L$ of training samples increases, $\rhob$ decreases as $1/\sqrt{L}$. On the other hand, $\chib$ is constant with respect to $L$ and provides a floor for the bound.
Fig.~\ref{fig:CAOL:anal} illustrates $\chib$ for CAOL iterates from different image datasets, and provides empirical evidence that this term can indeed be small in real data.
If the model mismatch is sufficiently uncorrelated with the training samples,
i.e., $\chib$ is practically zero, then
only the $\rhob$ term remains and this term decreases with $L$. Namely, if model mismatch is entirely uncorrelated with the training samples, then using many samples decreases the error bound to (effectively) zero.

\section{Related Works} \label{sec:rel}

Sample complexity \cite{Bajwa&etal:bookCh} and synthesis error \cite{Singh&Poczos&Ma:18PMLR} have been studied in the context of synthesis operator learning (e.g., dictionary learning \cite{Aharon&Elad&Bruckstein:06TSP});
see the cited papers and references therein.
A similar understanding for (C)AOL has however remained largely open; existing works focus primarily on establishing (C)AOL models and their algorithmic challenges \cite{Yaghoobi&etal:13TSP, Hawe&Kleinsteuber&Diepold:13TIP, Cai&etal:14ACHA, Ravishankar&Bressler:15TSP, Chun&Fessler:18arXiv}.
The authors in \cite{Seibert&etal:16TSP} studied sample complexity for a patch-based AOL method, but the form of their model differs from that of ours \R{sys:CAOL:orth}. Specifically, they consider the following AOL problem:
$\min_{D} \sum_l f(D^T \hat{x}_l) + g(D)$,
where $f(\cdot)$ is a sparsity promoting function (e.g., a smooth approximation of the $\ell^0$-quasi-norm \cite{Seibert&etal:16TSP}),
$g(\cdot)$ is a regularizer or constraint for the filter matrix $D$,
and $\{ \hat{x}_l : l = 1,\ldots,L \}$ is a set of training {\em patches} (not images).

\section{Proof of Theorem~\ref{t:det}}

Rewriting \R{opt:filter:orth} yields that
$\Ds$ is a solution of the (scaled)
orthogonal Procrustes problem~\cite[\S S.\Romnum{7}]{Chun&Fessler:18arXiv}:
\be{
\label{opt:filter:orth:procrustes}
\argmin_{D} \| \widetilde{\Psi} D - \widetilde{\ZZ} \|_F^2,
\quad \mbox{subj.~to} ~ D D^H = \frac{1}{R} \cdot I,
}
where $\widetilde{\Psi} \in \bbC^{L N \times R}$
arises by stacking $\Psi_1,\dots,\Psi_L$ vertically
and $\widetilde{\ZZ} \in \bbC^{L N \times K}$
arises likewise from $\Z{1},\dots,\Z{L}$.
Similarly, since $\Psi_l \Dt = \Zt{l}$ as in \R{sys:filter},
$\Dt$ is a solution of the analogous (scaled)
orthogonal Procrustes problem
\be{
\label{opt:filter:orth:procrustes:true}
\argmin_{D} \| \widetilde{\Psi} D - \widetilde{\ZZ}_{\textnormal{true}} \|_F^2,
\quad \mbox{subj.~to} ~ D D^H = \frac{1}{R} \cdot I,
}
where $\widetilde{\ZZ}_{\textnormal{true}} \in \bbC^{L N \times K}$
arises by stacking $\Zt{1},\dots,\Zt{L}$ vertically.

By assumption, both $\widetilde{\Psi}^H \widetilde{\ZZ}$ and $\widetilde{\Psi}^H \widetilde{\ZZ}_{\textnormal{true}}$ are full row rank
and so~\R{opt:filter:orth:procrustes} and~\R{opt:filter:orth:procrustes:true}
have unique solutions given by the unique (scaled) polar factors
\ea{
\Ds &= \frac{1}{\sqrt{R}} \Q(\widetilde{\ZZ}^H  \widetilde{\Psi})^H &
\Dt &= \frac{1}{\sqrt{R}} \Q(\widetilde{\ZZ}_{\textnormal{true}}^H \widetilde{\Psi})^H
}
where $\Q(\cdot)$ denotes the polar factor of its argument,
and can be computed as $\mbox{Q}(A) = W V^H$
from the (thin) singular value decomposition $A = W \Sigma V^H$.

Thus we have
\ea{
\label{eq1:t:det:prf}
&~\| \Ds - \Dt \|_F^2
\nn \\
&= \frac{1}{R} \| \Q(\widetilde{Z}_{\textnormal{true}}^H \widetilde{\Psi}) - \Q(\widetilde{\ZZ}^H  \widetilde{\Psi}) \|_F^2
\nn \\
&\leq
  \frac{1}{R} \| \widetilde{E}^H \widetilde{\Psi} \|_F^2
  \Bigg\{
    \bigg[
      \frac{2}{\svd{R}(\widetilde{Z}_{\textnormal{true}}^H \widetilde{\Psi}) + \svd{R}(\widetilde{\ZZ}^H \widetilde{\Psi})}
    \bigg]^2
    \nn \\
    &\qquad\qquad\qquad\quad
    +
    \bigg[
      \frac{1}{\max\{\svd{R}(\widetilde{Z}_{\textnormal{true}}^H \widetilde{\Psi}),\svd{R}(\widetilde{\ZZ}^H \widetilde{\Psi} )\}}
    \bigg]^2
  \Bigg\}
\nn \\
&\leq
\frac{1}{R} \| \widetilde{E}^H \widetilde{\Psi} \|_F^2 \Bigg\{
  \bigg[ \frac{2}{\svd{R}(\widetilde{Z}_{\textnormal{true}}^H \widetilde{\Psi})} \bigg]^2
  +
  \bigg[ \frac{1}{\svd{R}(\widetilde{Z}_{\textnormal{true}}^H \widetilde{\Psi})} \bigg]^2
\Bigg\}
\nn \\
&= \frac{5}{R} \frac{\| \widetilde{\Psi}^H \widetilde{E} \|_F^2}{\svd{R}^2(\widetilde{\Psi}^H \widetilde{Z}_{\textnormal{true}})}
= \frac{5}{R}
  \frac{\| \sum_{l=1}^L \Psi_l^H E_l \|_F^2}
    {\svd{R}^2(\sum_{l=1}^L \Psi_l^H \Zt{l})}
}
where $\widetilde{E} = \widetilde{\ZZ} - \widetilde{\ZZ}_{\textnormal{true}}$ is exactly $E_1,\dots,E_L$ stacked vertically,
and $\svd{r}(\cdot)$ denotes the $r$th largest singular value of its argument.
The first inequality holds by the perturbation bound in \cite[Thm.~3]{Li:95SIAM:JMAA},
and the second holds since $\svd{R}(\widetilde{Z}^H \widetilde{\Psi}) \geq 0$.
Recalling that $\Zt{l} = \Psi_l \Dt$,
we rewrite the denominator of \R{eq1:t:det:prf} as
\ea{
\label{eq2:t:det:prf}
&~\svd{R}^2\Big( \sum_{l=1}^L \Psi_l^H \Zt{l} \Big)
= \svd{R}^2\Big( \sum_{l=1}^L \Psi_l^H \Psi_l \Dt \Big)
\nn \\
&= \frac{1}{R} \svd{R}^2\Big( \sum_{l=1}^L \Psi_l^H \Psi_l \Big)
= \frac{1}{R} \eigmin^2\Big( \sum_{l=1}^L \Psi_l^H \Psi_l \Big)
,
}
where the second equality holds because $\Dt \Dt^H = (1/R)I$.
Substituting \R{eq2:t:det:prf} into \R{eq1:t:det:prf}
yields \R{t:det:bound:new}.

\section{Proof of Corollary~\ref{c:det:randNoise}}

Taking the expectation of \R{t:det:bound:new} over the model mismatch amounts to taking the expectation of the numerator of the upper bound in \R{t:det:bound:new}:
\ea{
\label{c:orth:pert:randNoise:prof1}
&~ \bbE \Big\| \sum_{l=1}^L \Psi_l^H E_l \Big\|_F^2
\nn \\
&= \sum_{l=1}^L \bbE \big\| \Psi_l^H E_l \big\|_F^2
= \sum_{l=1}^L \tr\Big(\Psi_l^H \bbE\{ E_l E_l^H\} \Psi_l\Big)
\nn \\
&\leq \sum_{l=1}^L \lambda_{\textnormal{max}}(\bbE\{ E_l E_l^H\}) \cdot \|\Psi_l\|_F^2
\leq \bar{\sigma}^2 \cdot \sum_{l=1}^L \|\Psi_l\|_F^2,
}
where the first equality holds by using the assumption that $E_l$ is zero-mean and independent over $l$,
the second equality follows by expanding the Frobenius norm then applying linearity of the trace and expectation,
the first inequality holds since $v^H M v \leq \lambda_{\textnormal{max}}(M) \cdot \|v\|_2^2$ for any vector $v$ and Hermitian matrix $M$,
and the last inequality follows from the definition of $\bar{\sigma}^2$.
Rewriting \R{c:orth:pert:randNoise:prof1} using the identity
$
\sum_{l=1}^L \|\Psi_l\|_F^2
= \sum_{l=1}^L \tr(\Psi_l^H\Psi_l)
= \tr ( \sum_{l=1}^L\Psi_l^H\Psi_l )
$
yields the result \R{c:det:randNoise:bound}.

\section{Proof of Theorem \ref{t:conc}}
\label{p:conc}

We derive two high probability bounds,
one each for the numerator and denominator of \R{t:det:bound:new}.
Then, the bound \R{t:det:bound:prob} with probability \R{t:det:prob}
follows by combining the two via a union bound.
Before we begin, note that \R{eq:conc:asbound}
implies that $\|\Psi\|_2 \leq \|\Psi\|_F \leq \gamma \sqrt{R}$
almost surely; our proofs use this inequality multiple times.

\subsection{Upper bound for numerator}

Observe first that
\ea{
\Big\| \sum_{l=1}^L \Psi_l^H E_l \Big\| _F
&=
\Big\|
  L \bbE(\Psi_l^H E_l)
  +
  \sum_{l=1}^L \{\Psi_l^H E_l-\bbE(\Psi_l^H E_l)\}
\Big\|_F
\nn \\
\label{conc:numer:decomp}
&\leq
L \| \bbE(\Psi_l^H E_l) \|_F
+
\Big\| \sum_{l=1}^L \xi_l \Big\|_2
,
}
where $\xi_l := \vect\{\Psi_l^H E_l-\bbE(\Psi_l^H E_l)\} \in \bbC^{RK}$ for $l = 1,\dots,L$.
We next bound $\| \sum_{l=1}^L \xi_l \|_2$ via the vector Bernstein inequality~\cite[Cor.~8.44]{Foucart&Rauhut:book}.
Note that $\xi_1,\dots,\xi_L$ are i.i.d.
with $\bbE \xi_l = 0$ (by construction).
Furthermore, $\xi_l$ is almost surely bounded as
\eas{
\|\xi_l\|_2 &= \|\Psi_l^H E_l-\bbE(\Psi_l^H E_l)\|_F \\
&\leq \|\Psi_l^H E_l\|_F + \|\bbE(\Psi_l^H E_l)\|_F
&&\text{(Triangle ineq.)} \\
&\leq \|\Psi_l^H E_l\|_F + \bbE\|\Psi_l^H E_l\|_F
&&\text{(Jensen's ineq.)} \\
&\leq \|\Psi_l\|_F\|E_l\|_F + \bbE\|\Psi_l\|_F\|E_l\|_F
& \\
&\leq 2\sigma\gamma\sqrt{R}
.
&
}
Thus the vector Bernstein inequality~\cite[Cor.~8.44]{Foucart&Rauhut:book}
yields that for any $t > 0$,
\be{
\label{conc:vector:bernstein:bound}
  \Big\| \sum_{l=1}^L \xi_l \Big\|_2
  \leq \sigma \sqrt{L} \sqrt{\tr(\Lambdab)} + t
,
}
with probability at least
\be{
\label{conc:vector:bernstein:prob}
1
-
\exp\bigg\{
  \frac{-t^2/2}{3L(2\sigma\gamma\sqrt{R})^2+t(2\sigma\gamma\sqrt{R})/3}
\bigg\}.
}
We obtained \R{conc:vector:bernstein:bound} by the following simplification:
\begingroup
\allowdisplaybreaks
\eas{
\bbE \Big\| \sum_{l=1}^L \xi_l \Big\|_2
&\leq \sqrt{\bbE \Big\| \sum_{l=1}^L \xi_l \Big\|_2^2}
= \sqrt{L\bbE \|\xi_l\|_2^2} \\
&= \sqrt{L\bbE \|\Psi_l^H E_l - \bbE(\Psi_l^H E_l)\|_F^2} \\
&=
\sqrt{L\{\bbE \|\Psi_l^H E_l\|_F^2 - \|\bbE(\Psi_l^H E_l)\|_F^2\}} \\
&\leq \sqrt{L\bbE \|\Psi_l^H E_l\|_F^2}
\leq \sqrt{L\bbE (\| \Psi_l \|_F^2 \| E_l \|_F^2)} \\
&\leq \sqrt{L \sigma^2 \bbE \| \Psi_l \|_F^2}
= \sigma\sqrt{L}\sqrt{\tr(\Lambdab)}
,
}
\endgroup
where the third equality holds by $\bbE \| A - \bbE A \|_F^2 = \sum_{i,j} \bbE ( A_{i,j} - \bbE A_{i,j} )^2 = \sum_{i,j} \bbE A_{i,j}^2 - ( \bbE A_{i,j} )^2 = \bbE \| A \|_F^2 - \| \bbE A \|_F^2$.
We obtained \R{conc:vector:bernstein:prob} by the following simplifications:
\begin{gather*}
\sup_{\|x\|_2 \leq 1} \bbE |x^H\xi_l|^2
\leq \bbE \|\xi_l\|_2^2 \leq (2\sigma\gamma\sqrt{R})^2
, \\
\bbE \Big\| \sum_{l=1}^L \xi_l \Big\|_2
\leq
\bbE \sum_{l=1}^L \|\xi_l\|_2
\leq L \bbE \|\xi_l\|_2
\leq L(2\sigma\gamma\sqrt{R})
.
\end{gather*}
Applying \R{conc:vector:bernstein:bound}
and \R{conc:vector:bernstein:prob}
with $t = 2 \sigma \gamma \sqrt{R} L\delta$
to the square of \R{conc:numer:decomp}
yields
\begingroup
\setlength{\thinmuskip}{1.5mu}
\setlength{\medmuskip}{2mu plus 1mu minus 2mu}
\setlength{\thickmuskip}{2.5mu plus 2.5mu}
\fontsize{9.5pt}{11.4pt}\selectfont
\be{
\label{conc:numer:bound}
\Big\| \sum_{l=1}^L \Psi_l^H E_l \Big\|_F^2 \\
\leq
L^2\Big\{
  \sigma\sqrt{\tr(\Lambdab)/L}
  +
  \| \bbE(\Psi_l^H E_l) \|_F
  +
  2 \sigma \gamma \sqrt{R} \delta
\Big\}^{\!\!2}
,
}
\endgroup
with probability at least $1 - \exp( -L \frac{\delta^2/2}{3+\delta/3} )$.

\subsection{Lower bound for denominator}

Observe that
$
\sum_{l=1}^L \Psi_l^H \Psi_l
= L \Lambdab + \sum_{l=1}^L \Lambda_l
$,
where $\Lambda_l := \Psi_l^H \Psi_l - \Lambdab$,
so Weyl's inequality \cite{Weyl:1912MA} yields
\be{
\label{eq:Cound:denom}
\eigmin \Big( \sum_{l=1}^L \Psi_l^H \Psi_l \Big)
\geq
\eigmin (L\bar{\Lambda})
-
\Big\| \sum_{l=1}^L \Lambda_l \Big\|_2
,
}
and it remains to bound $\| \sum_{l=1}^L \Lambda_l \|_2$.
We do so by using
the Matrix Bernstein inequality~\cite[Cor.~8.15]{Foucart&Rauhut:book}.

Note that $\Lambda_1,\dots,\Lambda_L$
are i.i.d. (since $x_1,\dots,x_L$ are i.i.d.)
and $\bbE \Lambda_l = 0$.
Furthermore, $\Lambda_l$ is almost surely bounded as
\eas{
\|\Lambda_l\|_2
&= \|\Psi_l^H \Psi_l - \bbE(\Psi_l^H \Psi_l)\|_2
\\
&\leq \|\Psi_l^H \Psi_l\|_2 + \|\bbE(\Psi_l^H \Psi_l)\|_2
&&\text{(Triangle ineq.)}
\\
&\leq \|\Psi_l^H \Psi_l\|_2 + \bbE \|\Psi_l^H \Psi_l\|_2
&&\text{(Jensen's ineq.)}
\\
&= \|\Psi_l\|_2^2 + \bbE \|\Psi_l\|_2^2
\leq 2\gamma^2R.
}
Thus, the Matrix Bernstein inequality~\cite[Cor.~8.15]{Foucart&Rauhut:book}
yields that for any $t > 0$,
\be{
\label{conc:bernstein}
\bbP \Big\{ \Big\| \sum_{l=1}^L \Lambda_l \Big\|_2 \geq t \Big\}
\leq 2 R \exp\Big\{\frac{-t^2/2}{L (2\gamma^2R)^2+2\gamma^2Rt/3}\Big\},
}
where we use the following simplification:
\bes{
\Big\| \sum_{l=1}^L \bbE \Lambda_l^2 \Big\|_2
= L \big\| \bbE \Lambda_l^2 \big\|_2
\leq L \bbE \|\Lambda_l\|_2^2
\leq L (2\gamma^2R)^2
.
}
Applying \R{conc:bernstein} with $t = 2\gamma^2RL\delta$
to the square of \R{eq:Cound:denom}
yields
\be{
\label{conc:eigenmin}
\eigmin^2\Big( \sum_{l=1}^L \Psi_l^H \Psi_l \Big)
\geq L^2 \{\eigmin(\Lambdab) - 2\gamma^2R \delta\}^2
,
}
with probability at least $1-2 R \exp (-L\frac{\delta^2/2}{1+\delta/3} )$.

\subsection{Combined bound}
Combining the bounds \R{conc:numer:bound} and \R{conc:eigenmin}
via a union bound yields \R{t:det:bound:prob}
with probability at least
\ea{
1
- \exp\bigg( -L\frac{\delta^2/2}{3+\delta/3} \bigg)
- 2 R \exp\bigg(-L\frac{\delta^2/2}{1+\delta/3}\bigg)
,
}
which is greater than or equal to \R{t:det:prob}.

\bibliographystyle{IEEEtran}
\balance
\bibliography{referenceBibs_Bobby}

\begin{thebibliography}{10}
\providecommand{\url}[1]{#1}
\csname url@samestyle\endcsname
\providecommand{\newblock}{\relax}
\providecommand{\bibinfo}[2]{#2}
\providecommand{\BIBentrySTDinterwordspacing}{\spaceskip=0pt\relax}
\providecommand{\BIBentryALTinterwordstretchfactor}{4}
\providecommand{\BIBentryALTinterwordspacing}{\spaceskip=\fontdimen2\font plus
\BIBentryALTinterwordstretchfactor\fontdimen3\font minus
  \fontdimen4\font\relax}
\providecommand{\BIBforeignlanguage}[2]{{%
\expandafter\ifx\csname l@#1\endcsname\relax
\typeout{** WARNING: IEEEtran.bst: No hyphenation pattern has been}%
\typeout{** loaded for the language `#1'. Using the pattern for}%
\typeout{** the default language instead.}%
\else
\language=\csname l@#1\endcsname
\fi
#2}}
\providecommand{\BIBdecl}{\relax}
\BIBdecl

\bibitem{Chun&Fessler:18arXiv}
\BIBentryALTinterwordspacing
I.~Y. Chun and J.~A. Fessler, ``Convolutional analysis operator learning:
  {A}cceleration and convergence,'' \emph{\emph{submitted}}, Jan. 2018.
  [Online]. Available: \url{http: //arxiv.org/abs/1802.05584}
\BIBentrySTDinterwordspacing

\bibitem{Chun&Fessler:18TIP}
------, ``Convolutional dictionary learning: Acceleration and convergence,''
  \emph{IEEE Trans. Image Process.}, vol.~27, no.~4, pp. 1697--1712, Apr. 2018.

\bibitem{Chun&Fessler:17SAMPTA}
------, ``Convergent convolutional dictionary learning using adaptive contrast
  enhancement ({CDL-ACE}): Application of {CDL} to image denoising,'' in
  \emph{Proc. Sampling Theory and Appl. (SampTA)}, Tallinn, Estonia, Jul. 2017,
  pp. 460--464.

\bibitem{Chun&Fessler:18Asilomar}
------, ``Convolutional analysis operator learning: {A}pplication to
  sparse-view {CT},'' in \emph{Proc. Asilomar Conf. on Signals, Syst., and
  Comput.}, Pacific Grove, CA, Oct. 2018, pp. 1631--1635.

\bibitem{Chun&etal:18arXiv:momnet}
I.~Y. Chun, Z.~Huang, H.~Lim, and J.~A. Fessler, ``{Momentum-Net}: {F}ast and
  convergent iterative neural network for inverse problems,''
  \emph{\emph{submitted}}, Jul. 2019.

\bibitem{Chun&Fessler:18IVMSP}
I.~Y. Chun and J.~A. Fessler, ``Deep {BCD}-net using identical
  encoding-decoding {CNN} structures for iterative image recovery,'' in
  \emph{Proc. IEEE IVMSP Workshop}, Zagori, Greece, Jun. 2018, pp. 1--5.

\bibitem{Chun&etal:18Allerton}
I.~Y. Chun, H.~Lim, Z.~Huang, and J.~A. Fessler, ``Fast and convergent
  iterative signal recovery using trained convolutional neural networkss,'' in
  \emph{Proc. Allerton Conf. on Commun., Control, and Comput.}, Allerton, IL,
  Oct. 2018, pp. 155--159.

\bibitem{Hastie&Tibshirani&Friedman:book}
T.~Hastie, R.~Tibshirani, and J.~Friedman, \emph{The elements of statistical
  learning: Data mining, inference, and prediction}, ser. Springer series in
  statistics.\hskip 1em plus 0.5em minus 0.4em\relax New York, NY: Springer,
  2009.

\bibitem{Mohri&Rostamizadeh&Talwalkar:book}
M.~Mohri, A.~Rostamizadeh, and A.~Talwalkar, \emph{Foundations of machine
  learning}.\hskip 1em plus 0.5em minus 0.4em\relax Cambridge, MA: MIT Press,
  2018.

\bibitem{Bajwa&etal:bookCh}
Z.~Shakeri, A.~D. Sarwate, and W.~U. Bajwa, ``Sample complexity bounds for
  dictionary learning from vector- and tensor-valued data,'' in
  \emph{Information Theoretic Methods in Data Science}, M.~Rodrigues and
  Y.~Eldar, Eds.\hskip 1em plus 0.5em minus 0.4em\relax Cambridge, UK:
  Cambridge University Press, 2019, ch.~5.

\bibitem{Singh&Poczos&Ma:18PMLR}
S.~Singh, B.~P{\'o}czos, and J.~Ma, ``Minimax reconstruction risk of
  convolutional sparse dictionary learning,'' in \emph{Proc. Int. Conf. on
  Artif. Int. and Stat.}, ser. Proc. Mach. Learn. Res., vol.~84, Playa Blanca,
  Lanzarote, Canary Islands, Apr. 2018, pp. 1327--1336.

\bibitem{Aharon&Elad&Bruckstein:06TSP}
M.~Aharon, M.~Elad, and A.~Bruckstein, ``\textit{K}-{SVD}: An algorithm for
  designing overcomplete dictionaries for sparse representation,'' \emph{IEEE
  Trans. Signal Process.}, vol.~54, no.~11, pp. 4311--4322, Nov. 2006.

\bibitem{Yaghoobi&etal:13TSP}
M.~Yaghoobi, S.~Nam, R.~Gribonval, and M.~E. Davies, ``Constrained overcomplete
  analysis operator learning for cosparse signal modelling,'' \emph{IEEE Trans.
  Signal Process.}, vol.~61, no.~9, pp. 2341--2355, Mar. 2013.

\bibitem{Hawe&Kleinsteuber&Diepold:13TIP}
S.~Hawe, M.~Kleinsteuber, and K.~Diepold, ``Analysis operator learning and its
  application to image reconstruction,'' \emph{IEEE Trans. Image Process.},
  vol.~22, no.~6, pp. 2138--2150, Jun. 2013.

\bibitem{Cai&etal:14ACHA}
J.-F. Cai, H.~Ji, Z.~Shen, and G.-B. Ye, ``Data-driven tight frame construction
  and image denoising,'' \emph{Appl. Comput. Harmon. Anal.}, vol.~37, no.~1,
  pp. 89--105, Oct. 2014.

\bibitem{Ravishankar&Bressler:15TSP}
S.~Ravishankar and Y.~Bresler, ``{$\ell_0$} sparsifying transform learning with
  efficient optimal updates and convergence guarantees,'' \emph{IEEE Trans.
  Sig. Process.}, vol.~63, no.~9, pp. 2389--2404, May 2015.

\bibitem{Seibert&etal:16TSP}
M.~Seibert, J.~W{\"o}rmann, R.~Gribonval, and M.~Kleinsteuber, ``Learning
  co-sparse analysis operators with separable structures,'' \emph{IEEE Trans.
  Signal Process.}, vol.~64, no.~1, pp. 120--130, Jan. 2016.

\bibitem{Li:95SIAM:JMAA}
R.-C. Li, ``New perturbation bounds for the unitary polar factor,'' \emph{SIAM
  J. Matrix Anal. Appl.}, vol.~16, no.~1, pp. 327--332, Jan. 1995.

\bibitem{Foucart&Rauhut:book}
S.~Foucart and H.~Rauhut, \emph{A mathematical introduction to compressive
  sensing}.\hskip 1em plus 0.5em minus 0.4em\relax New York, NY: Springer,
  2013.

\bibitem{Weyl:1912MA}
H.~Weyl, ``Das asymptotische verteilungsgesetz der eigenwerte linearer
  partieller differentialgleichungen (mit einer anwendung auf die theorie der
  hohlraumstrahlung),'' \emph{Mathematische Annalen}, vol.~71, no.~4, pp.
  441--479, Dec. 1912.

\end{thebibliography}

\end{document}